\documentclass{article}

\usepackage{neurips_2021}
\usepackage{amsmath}
\usepackage[utf8]{inputenc}
\usepackage[utf8]{inputenc} 
\usepackage[T1]{fontenc}    
\usepackage{hyperref}       
\usepackage{url}            
\usepackage{booktabs}       
\usepackage{amsfonts}       
\usepackage{nicefrac}       
\usepackage{microtype}      
\usepackage{xcolor}  
\usepackage[english]{babel}
 \usepackage{algorithm}
\usepackage{algpseudocode}
\usepackage{graphicx}
\usepackage{subcaption} 
\usepackage{float}
\usepackage{enumitem} 
\usepackage{authblk}

\usepackage{amsfonts}
\usepackage{times}
\usepackage{latexsym}
\usepackage{amssymb}
\usepackage{amsmath}
\usepackage{verbatim}
\newtheorem{theorem}{Theorem}

\newtheorem{corollary}[theorem]{Corollary}

\newtheorem{definition}[theorem]{Definition}

\newtheorem{lemma}[theorem]{Lemma}

\newtheorem{proposition}[theorem]{Proposition}

\newtheorem{remark}[theorem]{Remark}
\newenvironment{proof}{ \textbf{Proof:} }{ \hfill $\Box$}

\def\bb0{{\mathbb{0}}}

\def\opt{\mathsf{OPT}}

\def\bb{{\mathbf{b}}}

\def\b0{{\mathbf{0}}}

\def\bP{{\mathbf{P}}}

\def\b1{{\mathbf{1}}}

\def\bbD{{\mathbb{D}}}
\def\bbE{{\mathbb{E}}}

\def\bbP{{\mathbb{P}}}
\def\bbQ{{\mathbb{Q}}}
\def\bbR{{\mathbb{R}}}

\def\cA{\mathcal{A}}

\def\cE{\mathcal{E}}

\def\cI{\mathcal{I}}

\def\cM{\mathcal{M}}

\def\cO{\mathcal{O}}

\def\cR{\mathcal{R}}

\def\cU{\mathcal{U}}

\def\sfA{\mathsf{A}}

\def\sfr{{\mathsf{r}}}

\def\sf0{{\mathsf{0}}}

\def\nn{\nonumber}

\begin{document}
\title{Scheduling to Learn In An Unsupervised Online Streaming Model}

\author[1]{Rahul Vaze}
\author[2]{Santanu Rathod}
\affil[1]{TIFR, Mumbai}
\affil[2]{IIT-Bombay}

\maketitle
\begin{abstract}
An unsupervised online streaming model is considered where samples arrive in an online fashion over $T$ slots. There are $M$ classifiers, whose confusion matrices are unknown a priori. In each slot, at most one sample can be labeled by any classifier. The accuracy of a sample is a function of the set of labels obtained for it from various classifiers. The utility of a sample is a scalar multiple of its accuracy minus the response time (difference of the departure slot and the arrival slot), where the departure slot is also decided by the algorithm. 
Since each classifier can label at most one sample per slot, there is a tradeoff between obtaining a larger set of labels for a particular sample to  improve its accuracy, and its response time. The problem of maximizing the sum of the utilities of all samples is considered, where learning the confusion matrices, sample-classifier matching assignment, and sample departure slot decisions depend on each other. The proposed algorithm first learns the confusion matrices, and then uses a greedy algorithm for sample-classifier matching. A sample departs once its incremental utility turns non-positive. We show that the competitive ratio of the proposed algorithm is $\frac{1}{2}-{\mathcal O}\left(\frac{\log T}{T}\right)$.
\end{abstract}

\section{Introduction}
Dawid and Skene in what is now regarded as seminal work \cite{dawid1979maximum}, considered the basic problem of discovering ground truth or true labels of samples from multiple but possibly erroneous/noisy responses/measurements. The more modern paradigm considers the online variant of Dawid and Skene's model
for unsupervised learning (USL), where samples arrive over time and the problem is to discover their true labels using a large number of classifiers, typically
 `non-expert' crowdsourcing workers. 
With the advancement in cloud based services, this model has been successfully implemented in 
variety of platforms, such as Amazon Mechanical Turk.

Such systems, however, have limitations resulting from  the lack of knowledge of ground truth and the precise accuracy information of each of the classifiers. 
Thus, there is a need for aggregating algorithms that can infer the ground truth using a large ensemble of classifiers. There is a large 
body of work in this direction, where the main idea is to find a small set of good classifiers and/or estimating the accuracy of all classifiers from the obtained labels for different samples \cite{li2013error, raykar2010learning, shaham2016deep, karger2011iterative, ho2012online, 
natarajan2013learning, jordan, whitehill2009whose, liu2015online, gong2018incentivizing, nordio2018selecting, karger2014budget, 
sheng2008get}. 

To make the model formal, a confusion matrix $\bP$ is defined for each classifier, whose $i,j^{th}$ entry captures the probability that 
the classifier labels a sample as belonging to class $j$ when the true class label is $i$. Learning the confusion matrix for each classifier is part of the problem, using which the true label of a sample can be discovered with high probability. 

In this paper, we consider a competition model for the online USL in the Dawid-Skene model, where in each time slot, a set of samples arrive into the system. There are a total of 
$M$ classifiers, where  classifier $m$ is defined by its true confusion matrix $\bP_m$, that is unknown a priori. 
Without loss of generality, we consider that each classifier can label at most one sample in each time slot. See Remark \ref{rem:cap} for more details. The accuracy of a sample in a time slot  is defined as the probability of classifying it correctly using the optimal combining rule given its label set obtained till that time slot. 
 To consider a general model, we let that a sample can remain in system for more than one time slot to increase its accuracy in hope of getting more labels, however, at the cost of accruing delay. 
%
%
%
Thus, this model captures the tradeoff between sample accuracy and throughput of the system, where throughput is defined as the rate of sample departures.

The studied model is well-suited for real-time learning paradigms, such as image classification in social networks e.g. instagram, facebook, automated crowd controlling system, medical diagnostics, automated quality management in factories, large dataset handling applications in bioinformatics, etc., where not only good classification is needed but the speed/delay incurred matters. In the interest of space, we refer the reader to \cite{massoulie2016capacity, shah2020adaptive} for more discussion on applications.

\subsection{Related Work}
For USL models, the competition between accuracy and throughput has been studied in prior work \cite{massoulie2016capacity, shah2020adaptive, shakkott} by defining a metric called {\it capacity}. Capacity is defined to be the largest stochastic rate 
of arrival of samples per time slot, such that each sample can be guaranteed to have an accuracy above a certain threshold with high probability, 
while maintaining queue stability. 
Queue stability states that the long term average of the expected number of samples in the system (yet to depart the system) remains bounded. 
Even though the prior work \cite{massoulie2016capacity, shah2020adaptive,shakkott} captures the 
tension between accuracy of samples and their arrival rates, it only considers the metric of stability that is a long term throughput metric, and cannot account for more refined utilities such as per sample delay, etc.

To model a more fine grained system, in this work, we define a  per-sample utility, that is a function of the accuracy of that sample, 
and the time it takes to accrue that accuracy. Accuracy is directly related to the set of classifiers that label that sample. Thus, 
each sample would ideally like to get labels from the set of good or all classifiers, however, that requires waiting for multiple time slots because of the presence of other samples and each classifier labelling at most one sample per unit time, introducing a delay cost. In particular, we choose the sample utility to be a linear combination of its accuracy at its departure time, and the response time (departure time -arrival time), and the total utility is the sum of the utilities of all samples.

Thus, in each time slot, the problem is to find a matching between the outstanding samples and the classifiers, 
given the current estimates on the entries of the confusion matrices of each classifier, such that the overall utility is maximized. Even when the confusion matrices are perfectly known a priori, this problem is combinatorial and hard to solve. The problem becomes even more involved in the practical setting, since the quality of estimates of the confusion matrices depend on the prior matching decisions, while future matching decisions depend on the current estimates of the confusion matrices. 

To formulate the problem in the online setting, we use the metric of {\it competitive ratio} that is defined as the ratio of the utility of any algorithm and the utility of the optimal offline algorithm, minimized over all input sequences of samples. The optimal offline algorithm is assumed to have genie knowledge (perfect knowledge of the confusion matrices) and knows the entire input sequence non-causally. Thus, in contrast to \cite{massoulie2016capacity, shah2020adaptive,shakkott},  in this paper, we consider that the sample arrival sequence is arbitrary and not necessarily stochastic (e.g. Poisson).

\subsection{Our contributions}
\begin{itemize}[leftmargin=*]
\item With genie knowledge of the confusion matrices, we define a greedy algorithm and show that its competitive ratio is at least $\frac{1}{2}$. The main tool in deriving this result is to show that the utility of each sample is submodular, for which the greedy algorithm's competitive ratio is shown to be at least $\frac{1}{2}$ using results from recent work  \cite{nived} compared to classical work \cite{fisher1978analysis}. 

\item Next, we define a {\bf regret} metric between the greedy algorithm with genie knowledge and the greedy algorithm that has to learn the confusion matrices. We show that if the error between the estimate and the true value of the entries of the confusion matrices is small enough, then the matching decisions made by the greedy algorithm in the genie setting and the greedy algorithm after learning are identical. This allows us to show that the expected regret of the greedy algorithm that learns the confusion matrices is at most ${\mathcal O}(\log(T))$, where $T$ is the time horizon.
 
 \item Combining the above two results, we get our final result that the competitive ratio of the proposed greedy algorithm that also learns the confusion matrices is at least $\frac{1}{2}-{\mathcal O}\left(\frac{\log T}{T}\right)$.
 \item We also provide experimental results for both the synthetic and the real-world data sets to confirm our theoretical results.
 
\end{itemize}
\section{System Model}
We consider the online analogue of the generalized Dawid-Skene model \cite{dawid1979maximum} for unsupervised classification, similar to \cite{jordan, liu2015online, shakkott}.
We consider slotted time, where in each time slot/time $t$, a (new) set $A(t)$ of samples arrive. $S = \cup_{t\le T} A(t)$ is the complete set of samples that arrives over the total time horizon of $T$ slots. 
For each sample, the true label can take two values $\{\pm 1\}$, i.e., a sample can belong to 
one of two classes $C_1$ and $C_2$. The two class problem is considered for simplicity. All the results of the paper can be extended to multi-class problems with $K$ classes.

Let $[M] = \{1, \dots, M\}$. There are a total of $M$ classifiers, where the confusion matrix of the $m^{th}$ classifier is $\bP_m, m=1,\dots, M$, where 
$\bP_m(i,j)$ is the probability that the sample's true label is $i$ but is labeled as $j$ by classifier $m$.  
Following \cite{kantor}, we assume that classifier $m$ has a competence level $p_m$, which is the probability of making a correct prediction, regardless of
the original class. This implies that the confusion matrix  
$\bP_m(i,i)=\bP_m(j,j) = p_m$  for all $m, i\ne j$, i.e., $p_m$ is the probability that classifier $m$ labels correctly. 
The assumption that $\bP_m(i,i)=\bP_m(j,j) = p_m$ is primarily 
considered since the error probability bounds in closed form \cite{kantor} are available only for this case.
We also consider  that $ \frac{1}{2} < p_m < 1-\rho < 1$, i.e., each classifier is better than making a random guess while no classifier is perfect (similar to  \cite{jordan}).

\begin{remark}The considered model is equivalent to the one-coin model \cite{jordan} that is quite popular in the literature, where $\bP_m(i,i) = p_m$ and $\bP_m(i,j) = \frac{1-p_m}{K-1}$ for $K$ classes, $i\in \{1,\dots, K\}, i\ne j$. In this paper, we restrict to $K=2$.
\end{remark}
We let that the number of samples arriving in each slot $1\le |A(t)| \le  \eta$, where $\eta$ is a constant independent 
of $T$ and $M$.
Without loss of generality, we assume that each classifier can label at most one sample per time slot similar to \cite{shakkott}. Thus, in each time slot, at most $M$ samples can be labelled.
\begin{remark}\label{rem:cap}
If classifier $m$ can process $\gamma>1$ samples per time slot, then we make $\gamma$ copies of classifier $m$, each with the same confusion matrix $\bP_m$. Since all these $\gamma$ copies have the same confusion matrix  $\bP_m$, the rate at which   $\bP_m$ is learnt is accelerated by a factor of $\gamma$,  since the error bounds on learning the confusion matrix depend on the number of samples seen \cite{jordan}.  Thus with $\gamma>1$, the system gets a constant speed up both in terms of learning the confusion matrix $\bP_m$ and processing larger number of samples, however, does not change the order wise results we derive in the paper.
\end{remark}

We follow the model \cite{shakkott}, that the true label
and the individual classifiers' labels for each sample are generated once and fixed thereafter. Thus, repeatedly assigning a sample to the same classifier yields no benefit.

Let the time slot at which a sample $s$ {\it exits} the system be $t_{e,s}$. 
Sample $s$ exits the system on account of either being labelled by all the $M$ classifiers or if an algorithm $\sfA$ decides do so. 
We let new samples arrive at the beginning of the slot, while all exits happen at the end of any slot. Let the 
set of samples that exits the system at the end of slot $t$ be $E(t)$.

Let at the end of slot $t$, the set of outstanding samples that have arrived till the beginning of slot $t$ but have not yet exited the system be $O(t)$. Thus, 
$O(t) = O(t-1) \cup A(t) \backslash E(t)$.
For each sample $s\in O(t)$, let the set of classifiers from which it has already got a label by the end of slot $t$ be $M(s,t)\subseteq [M]$.
In slot $t+1$, for any 
sample $s\in O(t) \cup A(t+1)$, a label can be obtained from any one of the previously unused classifiers. In particular, in slot $t+1$, for $s\in O(t)$ from any classifier $m \notin M(s,t)$, while for $s\in A(t+1)$ from any classifier $m \in [M]$. 

For $m\in M(s,t)$, let $L_m(s) \in \{\pm 1\}$ be the label obtained from classifier $m$ for sample $s$. Given the (possibly partial) label set $M(s,t)$ ($M(s,t) \subset [M]$), the optimal decision rule (called the {\bf weighted majority}) \cite{liu2015online}
declares the label of $s$ to be $+1$ if 
\begin{equation}\label{eq:decrule}
\sum_{\substack{m \in M(s,t): \\ L_{m(s)} = +1}} \log \frac{p_m}{1-p_m} > \sum_{\substack{m \in M(s,t):\\ L_m(s)  = -1}} \log \frac{p_m}{1-p_m},
\end{equation}

and $-1$ otherwise.
 Let $\text{acc}_s(M(s,t))$ be the probability 
that the final label obtained using \eqref{eq:decrule}  with the label set $M(s,t)$ is in fact the true label.
With the weighted majority rule \eqref{eq:decrule},  from \cite{kantor}, 
\begin{equation}\label{eq:wmrlb}
1-\exp^{- \textsf{err}_{M(s,t)}} \le \text{acc}_s(M(s,t)) \le 1- \frac{3}{8\exp^{2\textsf{err}_{M(s,t)} + 4 \sqrt{\textsf{err}_{M(s,t)}}}}
\end{equation}  
where 
\begin{equation}\label{eq:err}
\textsf{err}_{M(s,t)} = \sum_{m\in M(s,t)} \textsf{err}_m,  \ \text{and} \ \textsf{err}_m= (p_m-1/2) \log \frac{p_m}{1-p_m}.
\end{equation}

Since the gap between the lower  and the upper bound \eqref{eq:wmrlb}  on the success probability is a constant, for the purposes of this paper, we let 
\begin{equation}\label{eq:acc}
\text{acc}_s(M(s,t)) =1- \exp^{-c \textsf{err}_{M(s,t)}}
\end{equation} for a constant $c$. We suppress the constant $c$ for the rest of the theoretical analysis for notational ease.
 
From \eqref{eq:acc}, since $p_m > 1/2$, each sample can improve its accuracy by getting labels from as many classifiers as possible. However, since at most $M$ samples can be labelled in each time slot, each sample can improve its accuracy only at the cost of delay by staying in the system for a longer time.   
\begin{definition} For each time slot $t$, any classifier $m$ can be assigned to any sample $s$. Thus, we consider the pair $(m,t)$ as the resource block $r_{m,t}$. Thus in each time slot $t$, there are $M$ resource blocks. Let the collection of all resource block across time horizon $T$ be denoted as $\cR = \{r_{m,t}: m=1,\dots, M,t=1,\dots, T\}.$
\end{definition}

\begin{definition}\label{defn:util}
For any $R\subseteq \cR$, the utility $u$ of sample $s$ is defined as  
\begin{equation}\label{eq:util}
u_s(R \cup \{x\}) = \max\{f_s(R), f_s(R\cup\{x\})\},
\end{equation}
where 
$$f_s(R) = w_s \text{acc}_s(R) - (D(R)-a_s),$$
 where $\text{acc}_s(R)$ is the probability 
that the label obtained using \eqref{eq:decrule} with label set from classifiers belonging to $R$ is in fact the true label, while $D(R)$ is the index of the latest time slot of any resource block in $R$, that counts the delay experienced by sample $s$ if all  classifiers of $R$ are used for it, where $a_s$ is the arrival time of the sample $s$, and $w_s>0$ is the sample weight that trades off the accuracy versus the delay cost.
\end{definition}

Note that expression $f_s$ is the `real' utility for sample $s$, however, it has the property that adding a new resource block $x$, $f_s$ can remain the same or decrease. This is on account of the classifier belonging to $x$ having being already used in $R$, thus providing no improvement in $\text{acc}_s$, or increase in delay because of inclusion of $x$ which increases $D$, thereby potentially decreasing $f_s$.
Since any reasonable algorithm will not assign a new resource block to sample $s$ even when it reduces its utility, the considered definition of utility $u_s(R\cup \{x\}) = \max\{f_s(R), f_s(R\cup\{x\})\}$ is natural,


Note that with this notation, if the final resource block subset assigned to sample $s$ is $R_s$, then the exit time $t_{e,s} =D(R_s)$ when the sample $s$ exits the system,  at which time, possibly $M(s,t_{e,s})\subset [M]$, i.e., sample $s$ is not labeled by all the classifiers. 

Since each classifier can label at most one sample per unit time, we can represent the decision about which sample should be labelled by which classifier at any time, as a bi-partite matching. In particular,  let $\cM_t$ be
a bi-partite matching between the set of outstanding samples $\{s: s \in O(t-1)\cup A(t)\}$ and the set of classifiers $[M]$ at time slot $t$, where an edge exists between $s\in O(t-1)\cup A(t)$ and $m\in [M]$ if $m \notin M(s,t-1)$, i.e., sample $s$ has not been labeled by classifier $m$ already. If an edge $e = (s,m)$ is part of matching $\cM_t$, then a label is obtained for sample $s$ from classifier $m$ at time slot $t$.

For any algorithm $\textsf{alg}$, the objective is to maximize the sum of the utilities \eqref{eq:util} across the samples, where the decision variables at the beginning of time slot $t$ are i)  matching $\cM_t$ between the samples and the classifiers, and ii) the exit time decision $t_{e,s}$ for each sample $s$. 
Note that since $p_m's$ are unknown, past matchings $\cM_i, i\le t$ of the algorithm controls the quality of the estimate of $p_m$ at time $t$, which consequently impacts the matching decision at time $t+1$, i.e., $\cM_{t+1}$.
We consider this problem in the online setup, where any algorithm $\textsf{alg}$ has only causal information, 
i.e., at time slot $t$, the algorithm does not know anything about the samples arriving in future (slots $> t$). Formally, the optimization problem is
\begin{equation}\label{defn:probsta}
\cU = \max_{\cM_t, t_{e,s}} \sum_{s} u_s,
\end{equation}
where decisions are allowed to use only causal information.

The performance metric for an online algorithm is called the competitive ratio, that is defined for an algorithm $\textsf{alg}$ as 
\begin{equation}\label{def:compratio}
\sfr_{\textsf{alg}} = \min_{\cI} \frac{ \cU_\cI ({\textsf{alg})}}{\cU_\cI(\opt)}, 
\end{equation}
where $\cI = \{A(t)\}$ is the input sequence of samples, and $\opt$ is the optimal offline algorithm (unknown) that knows the input sequence in advance, and the true value of $p_m's$ a priori. 
The goal is to design an algorithm with as large a competitive ratio as possible.

We next begin with some preliminaries, that will help us in analysis.

\begin{definition}\label{defn:submod}
Let $N$ be a finite set, and let $2^N$ be the power set of $N$. A real-valued set function $f:2^N\to\mathbb{R}$ is said to be {\it monotone} if $f(S) \le f(T)$ for $S \subseteq T \subseteq N$,  and {\it submodular} if  $
f(S \cup \{i\}) -  f(S)\ge  f(T \cup \{i\}) - f(T)$
for every $S\subseteq T\subset N$ and every $ i \notin T$. 
\end{definition}

We first show that the accuracy \eqref{defn:util} is a sub-modular function. 

\begin{proposition}\label{prop:submod} $\text{acc}_s$ \eqref{eq:acc} is a monotone and sub-modular set function.
\end{proposition}
\begin{proof} Checking monotonicity of $\text{acc}_s$ is straightforward since $p_m \ge 1/2$ for all $m$. 
Next, we check the  sub-modularity of $\text{acc}_s$. With slight abuse of notation, from \eqref{eq:acc}, the accuracy for sample $s$ with label set $V\subseteq [M]$ is denoted as
\begin{align}\label{eq:acc1}\text{acc}_s(V) &=1- \exp^{- \textsf{err}_{V}},
\end{align}
where $\textsf{err}_{V} = \sum_{m\in V} \textsf{err}_m$, 
and  $ \textsf{err}_m= (p_m-1/2) \log \frac{p_m}{1-p_m}$.
To check that $\text{acc}_s(t)$ is a sub-modular function in $M(s,t)$, we use definition \eqref{defn:submod}, and show that 
$\text{acc}_s(V \cup \{m\}) -  \text{acc}_s(V)\ge  \text{acc}_s(V'\cup \{m\}) - \text{acc}_s(V')$ where $m \notin V,V'$ and $V \subset V'\subset [M]$.
From \eqref{eq:acc1}, $ \text{acc}_s(V \cup \{m\}) -  \text{acc}_s(V)$
\begin{align}\nn
& =1- \exp^{-\textsf{err}_{V \cup \{m\}}} - \left(1- \exp^{- \textsf{err}_{V}}\right),\\ \label{eq:acc2}
 & = \exp^{- \textsf{err}_{V}} - \exp^{- \textsf{err}_{V \cup \{m\}}} = \exp^{- \textsf{err}_{V}}\left(1-\exp^{-\textsf{err}_m}\right).
\end{align}
Therefore, $\text{acc}_s(V \cup \{m\}) -  \text{acc}_s(V)\ge  \text{acc}_s(V'\cup \{m\}) - \text{acc}_s(V')$ where $m \notin V,V'$ and $V \subset V'\subset [M]$, is equivalent to showing 
$\exp^{- \textsf{err}_{V}}\left(1-\exp^{-\textsf{err}_m}\right) \ge \exp^{- \textsf{err}_{V'}}\left(1-\exp^{-\textsf{err}_m}\right)$, which is true 
since $V\subset V'$.
\end{proof}
\begin{proposition}\label{lem:submod} $f$ is a sub-modular set function.
\end{proposition}
Since $\text{acc}_s$ is sub-modular (Proposition \ref{lem:submod}), and $(D(R)-a_s)$ is a linear function and hence sub-modular, the proposition follows since linear combination of sub-modular functions is sub-modular.
\begin{lemma}
Utility function $u_s$ is a monotone and sub-modular set function.
\end{lemma}
\begin{proof}
For any $S,x \subset \cR$, by definition,  $u_s(S \cup \{x\})\ge u_s(S)$, thus monotonicity is immediate. For sub-modularity, let $S\subseteq T$ and $x\notin S,T$, then consider 
 $$u_s(T \cup \{x\})-u_s(T) =  \max\{f(T), f(T\cup\{x\})\} -  f(T).$$
 Case I: $f(T\cup\{x\})>f(T)$. In this case, we also get that $f(S\cup\{x\}) > f(S)$, since $\text{acc}_s(S)$ is sub-modular as shown in Proposition \ref{prop:submod}, and $D(S)$ is additive. Hence, $$u_s(S \cup \{x\})-u_s(S) =  \max\{f(S), f(S\cup\{x\})\} -  f(S) =  f(S\cup\{x\})\} -  f(S).$$  Since $f$ is sub-modular, we also have 
 $$f(S\cup\{x\})\} -  f(S) \ge f(T\cup\{x\}) - f(T).$$
 Thus, $u_s(T \cup \{x\})-u_s(T) = f(T\cup\{x\}) - f(T) \le f(T\cup\{x\}) - f(T) = u_s(S \cup \{x\})-u_s(S)$, proving sub-modularity of $u$.
 
 Case II $f(T\cup\{x\})\le f(T)$. In this case, $u_s(T \cup \{x\})-u_s(T)=0$, while by definition $u_s(S\cup \{x\})-u_s(S)\ge0$. Thus, again we get that $u_s(T \cup \{x\})-u_s(T)\le u_s(S \cup \{x\})-u_s(S)$ proving sub-modularity of $u$.
\end{proof}

Using these preliminaries, we next propose algorithms  and bound their competitive ratios.

\section{Algorithms}
\subsection{Genie Setting}
\begin{definition}
Let $\b1_{s,m, t}=1$ if sample $s$ is matched to resource block $r_{m,t}$, classifier $m$ in slot $t$ (called assignment), and zero otherwise. 
Any classifier cannot be matched to more than one sample in each slot, i.e., for each $m$, $\sum_{s\in O(t-1)\cup A(t)} \b1_{s,m,t}\le 1, \ \forall \ t $. 
Let the complete sample-classifier-slot assignment be defined as $\cA = 
\cup_{s,m,t}\{\b1_{s,m, t}\}$ and its restriction for sample $s$ be $\cA_s$.
\end{definition}

To define the algorithm, it is useful to write the increments of $\cU(\cA)$ as 
$\sigma(\b1_{s, m, t} |\cA) := \cU(\b1_{s, m, t} \cup \cA) -  \cU(\cA)$ obtained via incrementing the current solution $\cA$ with an additional assignment $\b1_{s,m,t}$ (matching sample $s$ with classifier $m$ in slot $t$) to completely describe the utility function $\cU$ in a compact form.

We propose a simple greedy algorithm (Algorithm \ref{alg:greedy-genie}), that on arrival of each new sample $s$, creates $M$ (equal to the number of classifiers) copies of that sample $s_1, \dots, s_M$. 
To model the restriction that a sample can be labeled by a classifier at most once, we enforce a constraint that any $k\le M$ copies of any sample have to be assigned 
to $k$ distinct classifiers, while the copies themselves are indistinguishable. For example, if $M=2$ with classifiers $m_1$ and $m_2$, the two copies of sample $s$ are $s_1, s_2$. Then $(s_1, m_1), (s_2, m_2)$ or $(s_1, m_2) (s_2, m_1)$ are the only two valid sample-classifier assignments possible, which could be done over different slots.

For sample $s^j$, at time $t$, let the set $M(s^j,t)\subseteq [M]$ be the set of classifiers for which at least one copy of $s^j$ has already been assigned to some $m\in M(s^j,t)$ at of before time $t$. Thus, $M(s^j,t)$ is the set of ineligible classifiers for any copy of sample $s^j$ in future. 
A classifier $m$ is defined to be {\it eligible} for sample $s^j$ at time $t$ if $m \notin M(s^j,t)$.

A classifier $m$ is defined to be {\it free} at time slot $t$, if there is no sample that has been assigned to it at time slot $t$ so far. 
Let the set of free classifiers at time $t$ be $F(t)\subseteq [M]$, where an element $f \in F(t)$ is $f = m$. 

With the knowledge of $p_m$, the proposed greedy algorithm (Algorithm \ref{alg:greedy-genie}) orders the classifiers in decreasing order of their accuracies $p_m$. For each time slot $t$, the algorithm picks a free classifier (in order) and assigns that sample copy to it, among the outstanding ones that are eligible,  that 
maximizes the incremental utility $\sigma(\b1_{s, m, t} |\cA)$ given prior assignment $\cA$, {\it as long as the increment $\sigma(\b1_{s, m, t} |\cA)$ is positive}. 

A sample $s^j$ {\it exits} the system as soon as its incremental valuation over all its outstanding copies turns non-positive over all possible eligible classifiers in the current and the future time slots. Note that if the incremental valuation of
assigning a sample copy $s^j_i$ to classifier $m$ is non-positive at time slot $t$, then it remains non-positive for  any time $t'\ge t$. So the exit decision can be computed efficiently.




\begin{algorithm}
\caption{Greedy Algorithm} \label{alg:greedy-genie}
\begin{algorithmic}[1]
\State \textbf{Input} Confusion Matrices $\bP_m, m \in [M]$
    \State \textbf{Initialize:} $\cA= \emptyset, O(0) = \emptyset$, $i=0$, 
    
    \For{$t=1:  T$}
    \State Set of free classifiers $F(t) = \{m\in [M]\}$ for time slot $t$
     \State New samples $s \in A(t)$ arrive, make $M$ copies $s_1, \dots, s_M$ for each $s \in A(t)$ 
      \State $B(t) = \{ s^i_1 \dots s^i_M, s^i \in A(t) \}\cup O(t-1)$ \quad  \% $B(t)$ is the set of outstanding sample copies. 
             \State For sample $s^j$, $M(s^j,t) \subseteq [M]$ set of classifiers for which at least one copy of $s^j$ has been assigned to it
   
      \For {$m\in F(t)$ in decreasing order of $p_m$}
         \State Set of eligible sample copies assignable to classifier $m$, $Z_m = \{s^j_k \in B(t) | m \notin M(s^j,t)\}$

\If{$\max_{s^k_j\in Z_m} \ \sigma(\b1_{\{s^k_j, m, t\}} |\cA)\le 0$}
\State Break; 
 \Else
    \State{Allocate copy $s^\ell_r \in Z_m$ to  classifier $m$ where $s^\ell_r= \underset{{s^k_j\in Z_m}}{\text{argmax}} \ \sigma(\b1_{\{s^k_j, m, t\}} |\cA)$} 
    \State  Tie: broken arbitrarily
    \State $\cA \gets \cA \cup \{s^\ell_r, m, t\}$, $F(t) \gets F(t) \backslash \{m\}$ where $s^\ell_r$ for some $r$ is chosen in Step 16.

    \State $M(\ell, t') \gets M(\ell,t' ) \cup \{m\}, \forall t'\ge t$     
    \State $B(t) \gets B(t) \backslash \{s^k_j\}$, $O(t) \gets B(t)$
    \EndIf
 \EndFor
\State  All sample copies $s^j_k \in O(t)$ {\bf exit} for which $\max_{s^k_j\in  Z_m, m\in [M], t'\ge t} \sigma(\b1_{\{s^k_j, m, t'\}} |\cA) \le 0$

    \EndFor
    \State \textbf{Return} $\cA$
\end{algorithmic}

\end{algorithm}

%
%
%
%
%

\vspace{-0.1in}
\begin{theorem} \label{theorem:12apx}
For Problem \ref{defn:probstagen}, let the online assignment generated by Algorithm \ref{alg:greedy-genie} (with true value of $p_m$ known to it) be $\cA^g$ and the $\opt$ assignment be 
$\cA^\opt$, then we have that:
$\cU_\cI(\cA^g)\ge \frac{ \cU_\cI(\cA^\opt)}{2}$ for any input $\cI = \{A(t)\}$.
\end{theorem}
%
\vspace{-0.1in}
The above description is complete for Algorithm \ref{alg:greedy-genie} as long as it knows the exact values of the entries of the confusion matrices $p_m$. In practice, they are unknown and need to be learnt. 
Suppose we use Algorithm \ref{alg:greedy-genie} in the realistic setting when a learning module is used to estimate $p_m$. In this case, the estimate of $p_m$ at time $t$ depends on the history of the matching decisions, which in turn depends on the prior estimate of $p_m$. 
This joint learning and matching aspect  makes the problem interesting. 

Next, to handle this joint learning and matching aspect, we define a {\bf regret} metric, that will compare the performance of Algorithm \ref{alg:greedy-genie} with genie access 
(true knowledge of $p_m$), and another matching algorithm $\sfA$, where the value of $p_m$ is estimated from prior matching information, and then used for matching. 

For an algorithm $\sfA$, let ${\hat p}_m(t)$ be the estimated value of the true $p_m$ at time $t$. 
For the genie setting, ${\hat p}_m(t) = p_m$ for all $t$.
Let the output (assignments made) at time $t$ with Algorithm \ref{alg:greedy-genie} be $\cA^g(t)$  that uses true value of $p_m$, while the output of any Algorithm $\sfA$ at time $t$ be ${\hat \cA}(t)$ that uses ${\hat p}_m(t)$. 
Then the regret for algorithm $\sfA$ is defined as 
\begin{equation}\label{defn:regret}
\text{Regret}(\sfA, T)=\sum_{t=1}^T \cU(\cA^g(t)) - \cU({\hat \cA}(t)),
\end{equation}
where the total time horizon is $T$. Our goal is to show that the $\text{Regret}(\sfA, T) = o(T)$ for a certain learning algorithm $\sfA$ that 
is obtained by prefixing a learning component with Algorithm \ref{alg:greedy-genie}.



Towards that end, next, we obtain a structural result about Algorithm \ref{alg:greedy-genie} in Lemma \ref{lem:connection}. 
For a sample $s$,  let $s_1, \dots, s_M$ be its $M$ copies which can be assigned to $M$ distinct classifiers. Since the $M$ copies of a sample are indistinguishable, without loss of generality, we let the sample copies are assigned in order of their index, i.e., $s_i$ is assigned before $s_j$ if $i < j$.
Let at time $t$, $s_{k(t)}$ be the copy of sample $s$ with the largest index that has already been assigned to some classifier $m$ at time $t_{k(t)}\le t$, and the set of classifiers that have been matched to the $k(t)$ copies of sample $s$, $s_1 \dots s_{k(t)}$, be $M(s,t)$.

Using \eqref{eq:err}, the incremental utility  of assigning next sample copy $s_{k(t)+1}$ to a free classifier that is also eligible $m' \notin M(s,t')$ at time $t'\ge t$ is 
$\rho_{s,m',t'} = \left( w_{s} \exp^{-\textsf{err}_{M(s,t')}}\left(1 - \exp^{\textsf{err}_{m'}}\right) - (t'-t_{k(t)})\right).
$
Consider the difference of incremental utility $\rho_{s,m',t'}$ of assigning a copy $s^i_{k(t)}$ of sample $s^i$, and $s^j_{k(t)}$ of sample $s^j$, to some free and eligible classifier $m'$ at time $t'\ge t$, $m' \notin M(s^i, t') \cap M(s^j, t')$ and call it 
$\Delta_{ij}  = \rho_{s^i,m',t'}  - \rho_{s^j,m',t'} $.
Let 
$\Delta  = \min_X \Delta_{i,j}$, 
where set $X = \{i\ne j, M(s^i, t'),M(s^j, t') \subseteq [M], m \notin M(s^i, t') \cap M(s^j, t'), w_{s^i}, w_{s^k}\}$, all pairs of contending sample copies. $\Delta$ is the minimum gap in the difference of incremental utility of any two copies belonging to two distinct samples. 
 This definition is similar to the difference of the expected values of the top two arms in the multi-arm bandit settings \cite{auer2002finite}, that controls the regret.

 
Let the output (assignments made) of Algorithm \ref{alg:greedy-genie} with input $p_m$ and ${\hat p}_m$ be $\cA^g$ and ${\hat \cA}^g$, respectively. 
In the next Lemma, we show that if $\hat{\textsf{err}}_m$  (\eqref{eq:err} with $p_m = {\hat p}_m$), is close enough to $\textsf{err}_m$ for each classifier $m$, in terms of $\Delta$, then the assignments $\cA^g$ and ${\hat \cA}^g$ are identical, 
\begin{lemma}\label{lem:connection}
Let for the estimated value of $\hat{\textsf{err}}_m$  (\eqref{eq:err} with $p_m = {\hat p}_m$), the error bound
be $|\hat{\textsf{err}}_m- \textsf{err}_m| \le \Delta/(6w_{\max}2^M)$, where $w_{\max} \ge w_s \ \forall \ s$, an upper bound on the weight of any sample. 
Then $\cA^g={\hat \cA}^g$, and consequently the utilities of $\cA^g$ and ${\hat \cA}^g$ are the same. 
\end{lemma} 
The way to think about this result is that Algorithm \ref{alg:greedy-genie} is indifferent to errors in the input $p_m$ as long as they are `sufficiently' small. Lemma \ref{lem:connection} is a non-trivial result, since intuitively an algorithm need not satisfy this property. The proof crucially depends on the greedy nature (on incremental utility) of assignment decisions by Algorithm \ref{alg:greedy-genie}.
We are going to use Lemma \ref{lem:connection} to connect the decisions and utility of (genie) Algorithm \ref{alg:greedy-genie} and
(realistic) Algorithm \ref{alg:realgreedy} (proposed next) since effectively they only differ in their input about the values of $p_m$, similar to Lemma \ref{lem:connection}. 
\vspace{-.2in}
\subsection{Greedy Algorithm with Learning}
\vspace{-.35in}
\begin{minipage}{.5\textwidth} 
Consider the following greedy matching algorithm (Algorithm \ref{alg:realgreedy}), whose operation is divided in two phases. The first phase  is dedicated for pure learning (used to learn $p_m$), where procedure Online  Learn is executed for time interval $[0, T_L]$. In the learning phase, for each $t\le T_L$ one sample from the arriving set $A(t)$ is randomly chosen for learning, and the remaining samples exit without being labelled. 
The second phase is dedicated for greedy matching (follows greedy algorithm (Algorithm \ref{alg:greedy-genie})) assuming the the learnt values of 
$p_m$ in the first phase are in fact the true values. 

\end{minipage}
\begin{minipage}{.5\textwidth} 
\begin{algorithm}[H]
\caption{Greedy Algorithm With Learning} \label{alg:realgreedy}
\begin{algorithmic}[1]
\State Phase I - \% Learning
\State Set of samples $S_L=\emptyset$ to be sent for learning 
\For {$t=1: T_L$}
\State set of samples $A(t)$ arrive
\State Randomly select one sample $s\in A(t)$
\State $S_L= S_L \cup \{s\}$
\EndFor
\State \{${\hat \bP}_m, m \in [M]\} =$ Online  Learn ($S_L$)
\State Phase II - \% Matching
\While {$T_L < t\le  T$}
\State Follow Algorithm \ref{alg:greedy-genie} with ${\hat \bP}_m$ to get $\cA$
\EndWhile
 \State \textbf{Return} $\cA$
\end{algorithmic}
\end{algorithm}
\end{minipage}


\begin{minipage}{.5\textwidth} %
We next describe the learning algorithm {\bf Online Learn} for our setup (also called the one coin model) that is the same as Algorithm 2 \cite{jordan} using the following definitions. Let the set of all arriving samples be $S$ and the number of classes be $K$. In this paper, we consider only $K=2$. For any sample $s \in S$, $L_m(s)$ be the label 
obtained from classifier $m$. For any two classifiers $m,m'$, let 
$N_{mm'} = \frac{K-1}{K}\left(\frac{\sum_{s=1}^{|S|} \b1_{L_m(s)=L_{m'}(s)}}{|S|}-\frac{1}{K}\right)$ and for any classifier $i$, $m$ and $m'$, $(m_i,m'_i) = 
\arg \max_{(m,m')}\{|N_{mm'}| : m\ne m'\ne i\}$. Let $\b1_{L_m(s)= k}$ if the $m^{th}$ classifier's label for sample $s$ is $k$. The {\bf main Theorem of this paper} is as follows. 
\end{minipage} %
\begin{minipage}{.5\textwidth} %

\begin{algorithm}[H]
\caption{Online Learn} \label{alg:learn}
\begin{algorithmic}[1]
\State Input : Observed labels $L_m(s), m\in [M], s\in S$, number of classes $K$ 
\State Initialize  ${\hat p}_i = \frac{1}{K} + \text{sign}(N_{i m_1})\sqrt{\frac{N_{i m_i}N_{im'_{i}}}{N_{m_i m'_i}}}$
\State If $\frac{1}{M}\sum_{i=1}^M {\hat p}_i\ge 1/K$, then $ {\hat p}_i =\frac{2}{K} -  {\hat p}_i$ for all $i\in [M]$
\State Iteratively execute the following three steps
\State \quad $q_{s k } \propto \exp(\sum_{i=1}^M\b1_{L_i(s)= k} \log ({\hat p}_i) $
\State \quad \quad \quad $ + \b1_{L_i(s)\ne k} \log \left(\frac{1-{\hat p}_i}{K-1}\right)))$
\State \quad Normalize $q_{s k }$ such that $\sum_{k=1}^K q_{s k } =1$
\State \quad $ {\hat p}_i = \frac{1}{|S|} \sum_{s=1}^{|S|} \sum_{k=1}^{K}q_{s k }\b1_{L_i(s)= k}$
\State Output $ {\hat p}_i, i\in [M]$ 
\end{algorithmic}
\end{algorithm}
\end{minipage}

\begin{theorem}\label{thm:main} Choosing $T_L = \Theta(\log T)$, the competitive ratio of Algorithm \ref{alg:realgreedy} is at least $\frac{1}{2}-{\mathcal O}\left(\frac{\log T}{T}\right)$.
\end{theorem}
\vspace{-0.2in}
\section{Experiments}
\vspace{-.15in}
In this section, we provide the empirical study comparing the competitive ratio and the regret between Algorithm \ref{alg:greedy-genie} and \ref{alg:realgreedy}. Ideally, we would like to include the optimal algorithm as well, however, since it is unknown, we have to preclude that. 
We perform three sets of experiments, first on synthetic dataset, and next two on real datasets, 
the binary Bird dataset \cite{bird} which has labels of bird species, and the multi-class DOG dataset \cite{dog} which has labels for the dog breeds. 

For all the experiments, the utility with Algorithm \ref{alg:greedy-genie} is computed assuming the true values of confusion matrices are available at the first slot, while for Algorithm \ref{alg:realgreedy}, the first $T_L$ slots are used for learning the confusions matrices, the utility is computed over the remaining $T-T_L$ slots using the learnt confusion matrices.

{\bf Synthetic Data:} We consider $M= 30$ classifiers/workers, and samples arrive over the time horizon $T$. We consider that each 
sample can belong to either of two classes $\{1,2\}$, and the true label of each sample is $\{1,2\}$ with equal probability. 
The number of incoming samples in a slot follows a Poisson distribution with rate $5$, and the sample weights $w_{s}$ are 
chosen uniformly randomly from the set $[3,4,\dots, 10]$.
For each classifier, the $2\times2$ confusion matrix is generated as follows: the diagonal entries $\bP_m(i,i)$ is given by $p_{m}= 0.9- 0.005 \times m$, $m=1,\dots, 30$, and the off-diagonal entries of $\bP_m$ are obtained as $1-p_m$. 

%

In Fig. \ref{fig:synthregret}, we plot the regret (normalized with the average number of samples arriving per time slot, and average weight of any sample to identify the scaling with $T$) and the competitive ratio between Algorithm \ref{alg:greedy-genie} and Algorithm \ref{alg:realgreedy}, as a function of time horizon $T$, where for each value of $T$, the best (numerically optimized) learning interval $T_L$ is found. Fig. \ref{fig:synthregret} shows that the simulated performance is better than the theoretical results; regret is $o(\log T)$ and the competitive ratio is at least $1/2-\log(T)/T$.
\begin{figure}
\centering
\includegraphics[width= .65\linewidth]{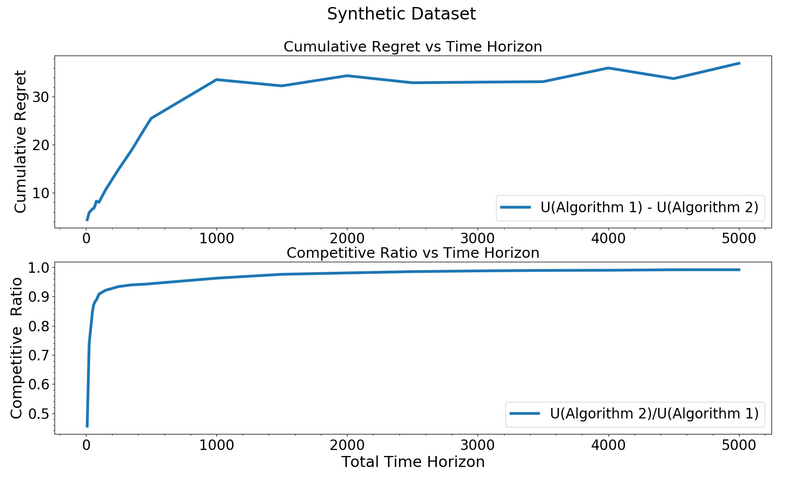}
\caption{Regret and competitive ratio for Algorithm \ref{alg:realgreedy} and Algorithm \ref{alg:greedy-genie} with  synthetic data}
\label{fig:synthregret}
\end{figure}
{\bf Real Data Sets:} Next, we use real two datasets, the Bird dataset \cite{bird} that has $M= 39$ classifiers/workers, and $|S|=108$ samples in total, and the DOG dataset \cite{dog} for which 
we take $M= 90$ classifiers and $|S|=798$ samples in total. 
We modified the multi-class DOG dataset to a binary dataset by clubbing classes 
$1,2$ as class $1$ and classes $3,4$ as class $2$. 
In Figs. \ref{fig:birdregret} and  \ref{fig:dogregret}, we plot the regret (normalized) and the competitive ratio between Algorithm \ref{alg:greedy-genie} and Algorithm \ref{alg:realgreedy}, as a function of time horizon $T$, where for each value of $T$, the best learning interval $T_L$ is found. More detailed results on best $T_L$ etc. can be found in the Appendix.

%
%


\begin{figure}[!h]
\centering
\includegraphics[width= 0.65\linewidth]{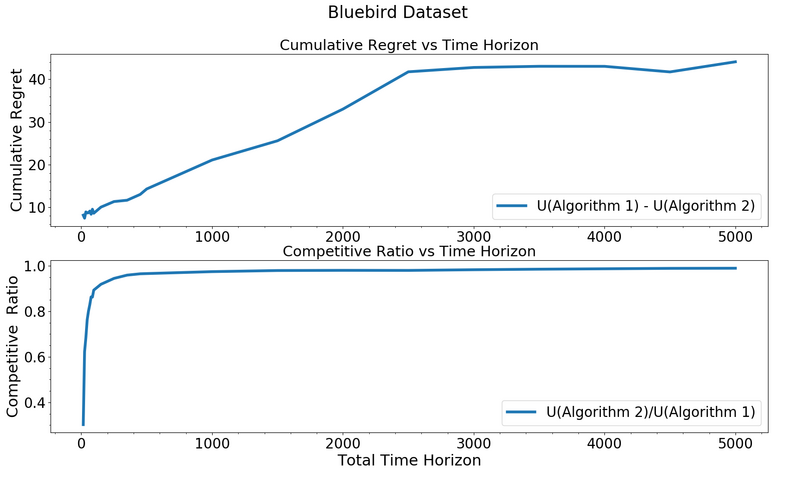}
\caption{Regret and competitive ratio for Algorithm \ref{alg:realgreedy} and Algorithm \ref{alg:greedy-genie} with  Bird data set }
\label{fig:birdregret}
\end{figure}

\begin{figure}[!h]
\centering
\includegraphics[height= 0.4\linewidth]{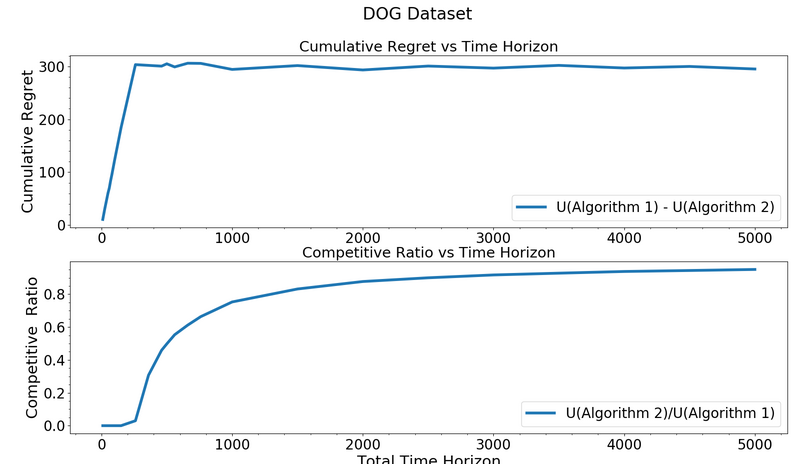}
\caption{Regret and competitive ratio for Algorithm \ref{alg:realgreedy} and Algorithm \ref{alg:greedy-genie} with  DOG data set }
\label{fig:dogregret}
\end{figure}


%
%
%
%
%
%



\section{Conclusions}
\vspace{-0.15in}
In this paper, we have considered a novel online competition model between samples in a unsupervised learning setup. Each arriving sample would like to 
get labelled by as many classifiers as possible to increase its accuracy while incurring the smallest delay. Each classifier, however, can label at most one sample per unit time slot, and this creates a tension between accuracy of a sample and the delay it incurs to achieve that accuracy. Each classifier is characterized by its confusion matrix, which has to be learnt as part of the problem. The problem is challenging since the matching decisions that assign samples to classifiers depends on the knowledge of the confusion matrix, and the quality of estimated confusion matrix depends on the past matching decisions.
We present a two phased algorithm, where in the first phase confusion matrices are learnt, which are then used by a greedy algorithm to assign samples to classifers that maximize the incremental utility. Using the submodularity of the utility function, we show that the competitive ratio of the proposed algorithm is close to $1/2$.
\bibliographystyle{unsrtnat}
\bibliography{refs}
\newpage
\section{Appendix}

\subsection{Proof of Theorem \ref{theorem:12apx}}
One can interpret the greedy algorithm (Algorithm \ref{alg:greedy-genie}) as where both the classifiers and the samples are arriving over time slots, and they are getting assigned sequentially in a  greedy manner. This aspect precludes the use of classical results on greedy algorithms  for standard online submodular maximization problems, to show that it is $\frac{1}{2}$-competitive for Problem \ref{defn:probstagen}. However, this case has been dealt in \cite{nived}, where a more general problem (Problem 3 \cite{nived}) has been considered for which a greedy algorithm (Algorithm \ref{alg:greedy-genie} is just a special case of that) has been shown to be at least $\frac{1}{2}$-competitive. 
Proof of Theorem \ref{theorem:12apx} is omitted since it directly follows from \cite{nived}.

\subsection{Proof of Lemma \ref{lem:connection}}\label{app:lem:connection}


Recall that the assignments made by Algorithm \ref{alg:greedy-genie}, with 
input $p_m$ and ${\hat p}_m$ are denoted as $\cA^g$, and ${\hat \cA}^g$, respectively.

Let at the end of time slot $t-1$, let $B(t-1)$ be the set of copies of samples that have arrived till time slot $t-1$, but not assigned to any classifier by the end of slot $t-1$, and have not exited. 
Then the set of copies of samples that are outstanding at the beginning of slot $t$ is $B(t) = B(t-1) \cup A_c(t)$, where 
$A_c(t) = \{s^i_{1}, \dots, s^i_{M} | s^i \in A(t)\}$, where $A(t)$ is the set of samples that arrives at the beginning of time slot $t$. The set $B(t)$ is updated after each iteration in which a sample copy is assigned to classifier in time slot $t$, i.e., if the $j^{th}$ copy of sample $s^i$, $s^i_{j}$ is assigned to some classifier, then   
$B(t) = B(t) \backslash \{s^i_{j}\}$.

Note that the $M$ copies of any sample are indistinguishable. Thus we let that they are assigned to classifiers in increasing order of their index, 
$s^i_{1}, \dots, s^i_{k},$ $k\le M$ for sample $s^i$.

We consider some iteration of the Algorithm \ref{alg:greedy-genie} that happens at time slot $t$, such that the assignments $\cA^g$ and ${\hat \cA}^g$ are identical until the previous iteration. Thus, the following holds. For a sample $s^i$, let $j_i$ be the smallest index for which its $j_i^{th}$ copy $s^i_{j_i}$ has not been assigned to any classifier, and $s^i_{j_i} \in B(t)$, $j_i\le M$. Thus, the $j_i^{th}$ copy of sample $s^i$ is part of the outstanding set of copies in the current iteration and all the previous $j_i-1$ copies of $s^i$ have already been assigned to some $j_i-1$ distinct classifiers in the past. 


Let $M(s^i_{j_i},t)$ be the 
(distinct) set of classifiers that  Algorithm \ref{alg:greedy-genie} or \ref{alg:realgreedy} (it is the same since they are have made identical assignment till now) has assigned the first $j_i-1$ copies, $s^i_{1}, \dots, s^i_{j_i-1}$ of sample $s^i$. Since we are interested in specific time slot $t$, for brevity, we drop the time index $t$ from $M(s^i_{j_i}, t)$, for the rest of the proof. Let $t_{i,j_i-1}\le t$ be the time slot in which 
the latest sample copy $s^i_{j_i-1}$ for sample $s^i$ is assigned to some classifier.

Recall that the incremental gain $\sigma_t(s^i_{j_i},m)$  in matching sample copy $s^i_{j_i}$ with classifier $m \notin M(s^i_{j_i})$ at time slot $t$, is $\sigma_t(s^i_{j_i},m)$
\begin{align}\nonumber
 &= w_s \left((1- \exp^{-\textsf{err}_{M(s^i_{j_i})\cup m}}) -(1- \exp^{-\textsf{err}_{M(s^i_{j_i})}}) \right)  - (t-t_{j_i-1}), \\\label{eq:incgain}
& = w_s \exp^{-\textsf{err}_{M(s^i_{j_i})}}\left(1 - \exp^{\textsf{err}_m}\right)-(t-t_{j_i-1}).
\end{align}
On any iteration, Algorithm \ref{alg:greedy-genie} matches classifiers $m\in [M]$ in decreasing order of $p_m$ (accuracy) to outstanding sample copies $s^i_{j_i} \in B(t)$
that maximizes the incremental gain $\sigma_t(s^i_{j},m)$. Note that for $\cA^g$, true value of $p_m$ and $\sigma_t(s_{ij},m)$ is used, while for ${\hat \cA}^g$    the estimated accuracy ${\hat p}_m$ is used to define ${\hat \sigma}_t(s_{ij},m)$.

Recall the definition of $\Delta$, i.e., $\Delta  = \min_X \Delta_{i,j}$, 
where set $X = \{i\ne j, M(s^i, t'),M(s^j, t') \subseteq [M], m \notin M(s^i, t') \cap M(s^j, t'), w_{s^i}, w_{s^k}\}$, all pairs of contending sample copies.

Property 1: Note that if $|\hat{\textsf{err}}_m(t)- \textsf{err}_m| <  \Delta/(2M)$, where $\Delta$, then 
${\hat p}_1(t)\ge \dots \ge {\hat p}_m(t)$ since $p_1\ge \dots \ge p_m$, i.e. the correct order of accuracy of classifiers is discovered with ${\hat p}_m(t)$. Since we assume that 
$|\hat{\textsf{err}}_m(t)- \textsf{err}_m| <  \Delta/(6 w_{\max} 2^M)$, clearly, $|\hat{\textsf{err}}_m(t)- \textsf{err}_m| <  \Delta/(2M)$.
 {\bf Thus, in any time slot $t$, 
the order in which classifiers (decreasing order of accuracy) are assigned by $\cA^g$ and ${\hat \cA}^g$  is the same.}

Case I : In the considered iteration of the Algorithm \ref{alg:greedy-genie}, let $m$ be the classifier that is going to be assigned  to some sample copy. If $B(t)$ (the set of outstanding samples in this iteration) consists of copies only
belonging to the same sample $s^i$, then trivially, in both $\cA^g$ and ${\hat \cA}^g$, sample copy $s^i_{j_i}$  is assigned to classifier $m$. Thus, $\cA^g$ and ${\hat \cA}^g$ remain the same, if they were identical before this iteration. 

Case II: Thus, 
the non-trivial case is when $B(t)$ consists of copies belonging to different samples. For any two distinct samples $s^i$ and $s^k$,  let $s^i_{j_i}, s^k_{j_k} \in B(t)$ be two copies belonging to different samples which are contending for classifier $m$. Next, we show that if classifier $m \in M(s^i_{j_i}) \cap M(s^k_{j_k})$ is assigned to $s^i_{j_i}$ in $\cA^g$,  so is the case in ${\hat \cA}^g$.  Since Algorithm \ref{alg:greedy-genie} assigns classifier $m \in M(s^i_{j_i}) \cap M(s^k_{j_k})$ at time slot $t$ to $s^i_{j_i}$, it means that 
\begin{equation}\label{eq:suffcondGg}
\sigma_t(s^i_{j_i},m) \ge \sigma_t(s^k_{j_k},m).
\end{equation}
Using \eqref{eq:incgain}, \eqref{eq:suffcondGg} we have that for $\cA^g$
\begin{align}\label{eq:dummy222}
w_{s^i} \exp^{-\textsf{err}_{M(s^i_{j_i})}}\left(1 - \exp^{\textsf{err}_m}\right)-(t-t_{j_i-1}) &  \ge  w_{s^k} \exp^{-\textsf{err}_{M(s^k_{j_k})}} 
\left(1 - \exp^{\textsf{err}_m}\right)
-(t-t_{j_k-1}).
\end{align} 
To claim the result, we intend to show that for ${\hat \cA}^g$,
${\hat \sigma}_t(s^i_{j_i},m) \ge {\hat \sigma}_t(s^k_{j_k},m)$, i.e.,
\begin{align}\nonumber
& w_{s^i} \exp^{-\hat{\textsf{err}}_{M(s^i_{j_i})}(t)}\left(1 - \exp^{\hat{\textsf{err}}_m(t)}\right)-(t-t_{j_i-1}) & \\ \label{eq:dummy223}
&  \ge  w_{s^k} \exp^{-\hat{\textsf{err}}_{M(s^k_{j_k})}(t)} 
\left(1 - \exp^{\hat{\textsf{err}}_m(t)}\right)
-(t-t_{j_k-1}).
\end{align}
where we have used the hypothesis of the Lemma, that the respective sets $M(s^i_{j_i})$ and $M(s^k_{j_k})$ are identical for both $\cA^g$ and ${\hat \cA}^g$ (since they have made identical decisions in past), and the only difference between \eqref{eq:dummy222} and \eqref{eq:dummy223} is in the value of $\hat{\textsf{err}}_V(t)$ and $\textsf{err}_V$ for $V \subseteq [M]$.

Recall  the definition of $\Delta$.
%
If \begin{equation}\label{eq:dummy992}
|\exp^{-\textsf{err}_{V}} - \exp^{-\hat{\textsf{err}}_{V}(t)}| \le \frac{\Delta}{3 .2 w_{\max} },
\end{equation} for any subset of classifiers $V \subseteq [M]$, where $w_s\le w_{\max}, \ \forall \ s$, then following simple algebraic steps (omitted due to lack of space), we have that \eqref{eq:dummy223} is true as long as \eqref{eq:dummy222} is true. 
 Consequently \eqref{eq:suffcondGg} holds for algorithm ${\hat \cA}^g$ as well.
Therefore, if $s^i_{j_i}$ is assigned to classifier $m$ in $\cA^g$ at time slot $t$, the same is true in ${\hat \cA}^g$ as well.

Next, we show how to satisfy \eqref{eq:dummy992}.
Note that $|V| \le M$ and $\textsf{err}_{V}$ \eqref{eq:err} is sum of at most $M$ terms, and, 
$\exp^{-\textsf{err}_{V}} = \prod_{m\in V}\exp^{-\textsf{err}_{m}}$ is a product of at most $M$ terms, where each term $\exp^{-\textsf{err}_{m}(t)} < 1$. Thus, if $|\exp^{-\textsf{err}_{m}} - \exp^{-\hat{\textsf{err}}_{m}(t)}| \le \frac{\Delta}{6 w_{\max}. 2^M }$, then it ensures that $|\exp^{-\textsf{err}_{V}} - \exp^{-\hat{\textsf{err}}_{V}(t)}| \le \frac{\Delta}{6 w_{\max}}$ for any $V \subseteq [M]$. Moreover, since $|\exp^{-|x|} - \exp^{-|y|}| \le |x-y|$ for all $x,y \in \bbR$,  
to ensure that $|\exp^{-\textsf{err}_{m}} - \exp^{-\hat{\textsf{err}}_{m}(t)}| \le \frac{\Delta}{6 w_{\max}2^M}$ it is sufficient that $|\textsf{err}_{m} -\hat{\textsf{err}}_{m}(t)| \le \frac{\Delta}{6 w_{\max}2^M}.$ Thus, \eqref{eq:dummy992} is true as long as $|\textsf{err}_{m} -\hat{\textsf{err}}_{m}(t)| \le \frac{\Delta}{6 w_{\max}2^M}.$
 Using similar arguments one can show that the exit time decisions also remains the same with $\cA^g$ and ${\hat \cA}^g$.

\subsection{Intermediate Results for the Proof of Theorem \ref{thm:main}}\label{app:lem:ptoerr}
We next state a result from \cite{jordan}, that will help us in upper bounding the regret \eqref{defn:regret} of Algorithm \ref{alg:realgreedy} in Lemma \ref{lem:regret} using Lemma \ref{lem:connection}.
Recall that $\bP_m = [\bP_{m,1} \dots \bP_{m,K}]$ is the confusion matrix of classifier $m$, with total $K$ classes and $\bP_{m,i}$ is the $i^{th}$ column of $\bP_m$ in the one-coin model in the symmetric case, where $\bP_m(i,i) = p_m$ and $\bP_m(i,j) = \frac{1-p_m}{K-1}$. 
Recall that we only consider $K=2$ in this paper. Let 
${\bar D} = \min_{\ell\ne \ell'}\frac{1}{M}\sum_{m=1}^M  \bbD_{\text{KL}}(\bP_{m,\ell}, \bP_{m,\ell^{'}})$ and $\bbD_{\text{KL}}(\bbP,\bbQ)$ is the KL-divergence between two discrete distributions $\bbP,\bbQ$. Let $\bar \kappa = \frac{1}{M}\sum_{m=1}^M\left(p_m-\frac{1}{K}\right) > 0$ and $\kappa_i$ is the $i^{th}$ largest element of vector $\{|p_m-\frac{1}{K}|\}_{m=1}^M$.
Lemma \ref{lem:jordan} provides an error bound on the entries of the confusion matrices $p_m$ with high probability as follows.
\begin{lemma}\label{lem:jordan}\cite{jordan} For Procedure 
Online Learn (part of Algorithm \ref{alg:realgreedy}), with $\alpha < p_m , \forall \ m$, for any $\delta > 0$ if the length of the learning phase is $$T_L = \Omega\left(\frac{\log \left(\frac{MK}{\delta}\right)}{
\left(\kappa_3^6\min\{{\bar \kappa}^2 \alpha^2 , (\alpha {\bar D})^2\}\right)}\right),$$ and the number of classifiers 
$$M = \Omega\left(\frac{ \log\left(\frac{1}{\alpha}\right)\log \left(\frac{T_LK}{\delta} + \log(T_L K)\right)}{{\bar D}}\right),$$ then $|\hat{\bP}_m(i,j)- \bP_m(i,j)| \le 2 \sqrt{\frac{\log \left(\frac{MK}{\delta}\right)}{T_L}}, \ \forall \ m,i,j$ with probability at least $1-\delta$.  \end{lemma}
Thus, if we want $|\hat{\bP}_m(i,j)- \bP_m(i,j)| \le \epsilon, \ \forall \ m , i, j$ with probability at least $1-\delta$, choosing $M\ge M^{\epsilon, \delta}$, and $T_L> f(\epsilon,\delta)$ for some explicit $f$, is sufficient from Lemma \ref{lem:jordan}\footnote{The exact expression for $f(\epsilon,\delta)$ is readily writable from Lemma \ref{lem:jordan}.}. Importantly, $T_L = \cO(\log (1/\delta))$.  
%
%
For applying Lemma \ref{lem:connection}, we need the bounds in terms of 
$\textsf{err}_{M(s,t)} = \sum_{m\in M(s,t)} \textsf{err}_m$, where $ \textsf{err}_m= (p_m-1/2) \log \frac{p_m}{1-p_m}$. Next, we translate bounds for $|\hat{\bP}_m(i,j)- \bP_m(i,j)| \le \epsilon, \ \forall \ m , i, j$ into a guarantee for $|\hat{\textsf{err}}_m- \textsf{err}_m| \le \epsilon, \ \forall \ m$.

\begin{lemma}\label{lem:ptoerr} With $M\ge M^{\epsilon, \delta}$, if the length of the learning phase is $T_L \ge  f\left(\epsilon/ {\left(\log(\dfrac{1-\rho}{\rho}) +\dfrac{1}{2}(\dfrac{1}{\rho}- \dfrac{1}{1-\rho})\right)}, \delta\right)$, then $|\hat{\textsf{err}}_m- \textsf{err}_m| \le \epsilon, \ \forall \ m$ with probability at least $1-\delta$ for any $\epsilon>0, \delta>0$, where recall that $\rho$ is such that $\frac{1}{2} \le p_m \le 1-\rho<1$.
\end{lemma}
\begin{proof} Recall that we have $p_{m} <1-\rho<1, \forall \ m \in [M]$ for some $\rho>0$. 
We want to show that if $| \hat{p}_m - p_m| \leq \epsilon$, then $| \textsf{err}_{m}- \hat{\textsf{err}_{m}}| \leq 
\left(\log(\dfrac{1-\rho}{\rho}) +\dfrac{1}{2}(\dfrac{1}{\rho}- \dfrac{1}{1-\rho})\right)\epsilon$.
Recall that $\textsf{err}(p)= (p - \dfrac{1}{2})\log(\dfrac{p}{1-p})$. Since $p_m \geq 0.5 \ \forall \ m$, we have that the first and second derivative of 
$\textsf{err}(p)$ for any $p \ge p_m$ is 
\begin{align}\label{eq:secondder1}
\textsf{err}^{'}(p)& = \log(\dfrac{p}{1-p}) + \dfrac{1}{2}\left(\dfrac{1}{1-p} - \dfrac{1}{p}\right) \geq 0, \\ \label{eq:secondder}
\textsf{err}^{''}(p)&= \dfrac{1}{p(1-p)} + \dfrac{1}{2}\left(\dfrac{1}{(1-p)^{2}} + \dfrac{1}{p^{2}}\right) \geq 0.
\end{align}

Consider pairs, $(p, \textsf{err}(p))$ and $(p-\epsilon, \textsf{err}(p - \epsilon))$  for any $\epsilon>0$. From \textit{\textbf{Lagrange's Mean Value Theorem (LMVT)}} we know that for a continuous and differentiable function $f(x)$, $\exists \ c \in (a,b)$ such that  
$\dfrac{f(b)- f(a)}{b -a}= f^{'}(c)$.
Applying \textit{LMVT} to function $\textsf{err}(p)$ on points $p -\epsilon$ and $p$ we get that $\exists \ c \in (p-\epsilon, p)$ such that, $ \dfrac{\textsf{err}(p) - \textsf{err}(p-\epsilon)}{p - (p - \epsilon)}  = \textsf{err}^{'}(c)$, and
\begin{align}\label{eq:dummyder1}
     \implies \textsf{err}(p) - \textsf{err}(p-\epsilon) & = \epsilon  \ \textsf{err}^{'}(c).
\end{align}
And since we know that $c \leq p \leq 1- \rho$ and $\textsf{err}^{''}(p) \geq 0$ from \eqref{eq:secondder}, we get that
\begin{equation}\label{eq:dummyder2}
    \textsf{err}^{'}(c) \leq \textsf{err}^{'}(p) \leq \textsf{err}^{'}(1- \rho).
\end{equation}

From \eqref{eq:dummyder1} and \eqref{eq:dummyder2}, we get that, $\textsf{err}(p)- \textsf{err}(p - \epsilon) \leq \epsilon \ \textsf{err}^{'}(1-\rho)$. 
From \eqref{eq:secondder1},  $\textsf{err}^{'}(1-\rho) \le \log\left(\dfrac{1-\rho}{\rho}\right) + \dfrac{1}{2}\left(\dfrac{1}{\rho} - \dfrac{1}{1-\rho}\right)$. Thus,  $|\textsf{err}(p)- \textsf{err}(p - \epsilon)| \le \left(\log\left(\dfrac{1-\rho}{\rho}\right) + \dfrac{1}{2}\left(\dfrac{1}{\rho} - \dfrac{1}{1-\rho}\right) \right) \epsilon.$
Therefore, if the error in estimating $p_m$ for any $m$ is at most $\epsilon$, i.e., $|{\hat p}_m - p_m|\le \epsilon$, then the error in estimating $| \textsf{err}_{m}- \hat{\textsf{err}_{m}}| \le 
 \left(\log\left(\dfrac{1-\rho}{\rho}\right) + \dfrac{1}{2}\left(\dfrac{1}{\rho} - \dfrac{1}{1-\rho}\right) \right) \epsilon$.
\end{proof}

\begin{corollary}\label{cor:relation} With $M\ge M^{\epsilon, \delta}$, for Algorithm \ref{alg:realgreedy} if $T_L = \Theta(\log T)$, then $|\hat{\textsf{err}}_m- \textsf{err}_m| \le \epsilon$ with probability at least $1-1/T^2$. 
\end{corollary}

Next, we bound the regret \eqref{defn:regret} of Algorithm \ref{alg:realgreedy} using Corollary \ref{cor:relation}, where the main idea is as follows. From Lemma 
\ref{lem:connection}, we know that Algorithm \ref{alg:greedy-genie} makes identical decisions, with input 
$p_m$ and ${\hat p}_m$, respectively, 
as long as $|\hat{\textsf{err}}_m- \textsf{err}_m| \le \Delta/(6w_{\max}2^M)$. This implies that the utility obtained by Algorithm \ref{alg:greedy-genie} and Algorithm \ref{alg:realgreedy} for the samples arriving in the matching phase of Algorithm \ref{alg:realgreedy} is identical as long as this inequality $|\hat{\textsf{err}}_m- \textsf{err}_m| \le \Delta/(6w_{\max}2^M)$ 
can be ensured at the end of learning phase of Algorithm \ref{alg:realgreedy}.  From Corollary \ref{cor:relation}, we get the length 
of the learning interval that is sufficient to ensure  
$|\hat{\textsf{err}}_m- \textsf{err}_m| \le \Delta/(6w_{\max}2^M)$ with high probability. Putting these two facts together we get the following crucial Lemma.
\begin{lemma}\label{lem:regret} If $T_L = \Theta(\log T)$ for Algorithm \ref{alg:realgreedy}, then the regret \eqref{defn:regret} of Algorithm \ref{alg:realgreedy} is at most $c\log T$ for some constant $c$ that does not depend on $T$, with probability at least $1-1/T^{2}$. 
\end{lemma}
Note that the constants in $\Theta(\log T)$ depend on $\Delta, \rho$, and $M$.

\begin{proof}
We know from Corollary \ref{cor:relation} that 
if the length of learning phase is $T_L = O( \log T)$ in Algorithm \ref{alg:realgreedy}, then event $\cE =\{ |\hat{\textsf{err}}_m- \textsf{err}_m| \le \Delta/(6.2^M w_{\max})\}$ happens with probability at least $1-1/T^2$.

Next, under the condition that $\cE$ is true, we compare the utility of Algorithm \ref{alg:greedy-genie} with genie access to $p_m$ and Algorithm \ref{alg:realgreedy}. Let the set of samples that arrives in the learning phase and the matching phase of Algorithm \ref{alg:realgreedy} be $S_1 = \cup_{t\le T_L} A(t)$ and $S_2=\cup_{t> T_L} A(t)$, respectively. 
Clearly, 
the total utility of the Algorithm \ref{alg:greedy-genie} with genie access to $p_m$ is upper bounded by the sum of the utility obtained from samples in $S_1$ and samples of $S_2$. With $\cA^g$ and ${\hat \cA}^g$ as the assignment of Algorithm \ref{alg:greedy-genie} and Algorithm \ref{alg:realgreedy}, respectively, we have, $\cU(\cA^g,T)$
\begin{align}\nn
 & \le  \sum_{t\in \ \text{learning phase} \ S_1}\cU(\cA^g(t))  + \sum_{t\in \ \text{learning phase} \ S_2}\cU(\cA(t)), \\ \nn
&\stackrel{(a)}   \le  \sum_{t\in \ \text{learning phase} \ S_1}\cU(\cA^g(t))  + \sum_{t\in \ \text{learning phase} \ S_2}\cU({\hat \cA}^g(t)), \\ \label{eq:finalregret} 
&\stackrel{(b)}   \le  c_1 \log T + \sum_{t\in \ \text{learning phase} \ S_2}\cU({\hat \cA}^g(t)),
\end{align}
where $(a)$ follows from Lemma \ref{lem:connection} since at the end of learning phase of Algorithm \ref{alg:realgreedy} $|\hat{\textsf{err}}_m(t)- \textsf{err}_m| \le \Delta/(6 w_{\max} 2^M)$, while $(b)$ follows from the fact that length of learning phase is $c \log T$, and in each time slot, a finite number of samples arrive, thus the maximum utility obtainable from samples arriving in learning phase is at most $c_1 \log T$ for some constant $c_1$.
Thus, \eqref{eq:finalregret} implies that the regret \eqref{defn:regret} of the greedy algorithm (Algorithm \ref{alg:realgreedy}) is at most $c_1 \log T$ as required with probability at least $1-1/T^2$.
\end{proof}
\begin{corollary}\label{cor:expregret}With $T_L = \Theta(\log T)$, the expected regret \eqref{defn:regret} of Algorithm \ref{alg:realgreedy} is at most $c\log T(1-1/T^2) + o(1/T)$.
\end{corollary}
From Lemma \ref{lem:regret}, we know that the regret of Algorithm \ref{alg:realgreedy} is at most $c\log T$  for some constant $c$ (that does not depend on $T$) with probability at least $1-1/T^{2}$. With the rest of the probability $1/T^2$, let the utility of Algorithm \ref{alg:realgreedy} be zero.
Note that the utility of the greedy algorithm with genie access to $p_m$ (Algorithm \ref{alg:greedy-genie})  is $\cO(T)$, since a constant number of samples arrive in each slot, and each 
samples' utility is bounded by a constant.  Thus, with probability $1/T^2$, the regret of Algorithm \ref{alg:realgreedy} is at most $\cO(T)$. Therefore the 
expected regret of  Algorithm \ref{alg:realgreedy} is at most $c\log T(1-1/T^2) + o(1/T)$.

\section{Proof of Theorem \ref{thm:main}}
\begin{proof} From Corollary \ref{cor:expregret}, we have that for Algorithm \ref{alg:realgreedy} with assignment ${\hat \cA}^g$, and Algorithm \ref{alg:greedy-genie} with assignment $\cA^g$, respectively,
$\bbE\{\cU_\cI({\hat \cA}^g)\} \ge \cU_\cI(\cA^g) - c \log T -o(1/T)$.
From Theorem \ref{theorem:12apx}, we know that  
$\cU_\cI(\cA^g)\ge \frac{ \cU_\cI(\opt)}{2}$. 
Combining these two results, we get that $\bbE\{\cU_\cI( {\hat \cA}^g)\}\ge \frac{ \cU_\cI(\opt)}{2} - c\log T  -o(1/T)$,
which is equivalent to 
\begin{align*}
\frac{\bbE\{\cU_\cI({\hat \cA}^g)\}}{\cU_\cI(\opt)}&\ge \frac{ 1}{2} - \frac{c\log T}{\cU_\cI(\opt)} -o\left(\frac{1}{T \cU_\cI(\opt)}\right), \\
& = \frac{1}{2} - {\mathcal O}\left(\frac{\log T}{T}\right) - o\left(\frac{1}{T^2}\right),
\end{align*}
since $\cU_\cI(\opt) = \Omega(T)$.
%

\end{proof}


\subsection{Detailed Experiments}
In this section, we provide the empirical study comparing the competitive ratio and the regret between Algorithm \ref{alg:greedy-genie} and \ref{alg:realgreedy}. Ideally, we would like to include the optimal algorithm as well, however, since it is unknown, we have to preclude that. 
We perform three sets of experiments, first on synthetic dataset, and next two on real datasets, 
the binary Bird dataset \cite{bird} which has labels of bird species, and the multi-class DOG dataset \cite{dog} which has labels for the dog breeds. 

For all the experiments, the utility with Algorithm \ref{alg:greedy-genie} is computed assuming the true values of confusion matrices are available at the first slot, while for Algorithm \ref{alg:realgreedy}, the first $T_L$ slots are used for learning the confusions matrices, the utility is computed over the remaining $T-T_L$ slots using the learnt confusion matrices.

{\bf Synthetic Data:} We consider $M= 30$ classifiers/workers, and samples arrive over the time horizon $T$. We consider that each 
sample can belong to either of two classes $\{1,2\}$, and the true label of each sample is $\{1,2\}$ with equal probability. 
The number of incoming samples in a slot follows a Poisson distribution with rate $5$, and the sample weights $w_{s}$ are 
chosen uniformly randomly from the set $[3,4,\dots, 10]$.
For each classifier, the $2\times2$ confusion matrix is generated as follows: the diagonal entries $\bP_m(i,i)$ is given by $p_{m}= 0.9- 0.005 \times m$, $m=1,\dots, 30$, and the off-diagonal entries of $\bP_m$ are obtained as $1-p_m$. 

\begin{figure}
\centering
\includegraphics[width= 0.8\linewidth]{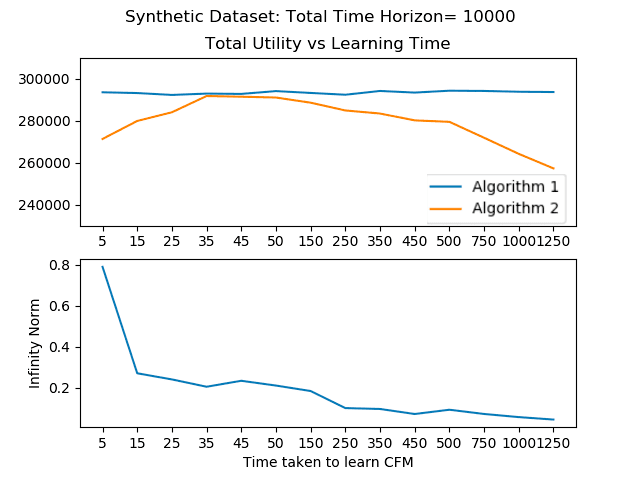}
\caption{Total Utility and Infinity norm as a function of learning time for synthetic data}
\label{fig:synth}
\end{figure}

In Fig. \ref{fig:synth}, for a fixed time horizon $T=10000$, we plot the accuracy ($\infty-norm$ of the error) in learning the confusion matrices with learning intervals of length $T_L \in \{5, 15, 25, 45, \cdots 1250\}$ using procedure Online Learn. We also plot the resulting utilities of Algorithm \ref{alg:greedy-genie} (that knows $p_m$ ahead of time) and Algorithm \ref{alg:realgreedy} (that works with learnt values of $p_m$ after $T_L$ slots),  in Fig. \ref{fig:synth}. Note that even though no learning is used for Algorithm \ref{alg:greedy-genie}, the utility curve is not constant because of inherent randomness in number of arriving samples in each slot, and their weights $w_s$.
%

In Fig. \ref{fig:synthregret}, we plot the regret (normalized with the average number of samples arriving per time slot, and average weight of any sample to identify the scaling with $T$) and the competitive ratio between Algorithm \ref{alg:greedy-genie} and Algorithm \ref{alg:realgreedy}, as a function of time horizon $T$, where for each value of $T$, the best learning interval $T_L$ is found. Fig. \ref{fig:synthregret} shows that the simulated performance is better than the theoretical results; regret is $o(\log T)$ and the competitive ratio is at least $1/2-\log(T)/T$.

\begin{figure}[h]
\centering
\includegraphics[width= .65\linewidth]{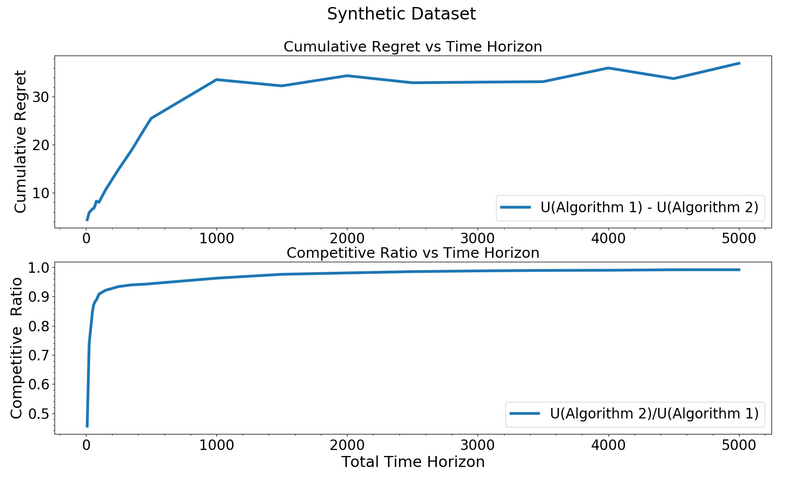}
\caption{Regret and competitive ratio for Algorithm \ref{alg:realgreedy} and Algorithm \ref{alg:greedy-genie} with  synthetic data}
\label{fig:synthregret}
\end{figure}

\begin{figure}
  \centering
  \includegraphics[width=.8\linewidth]{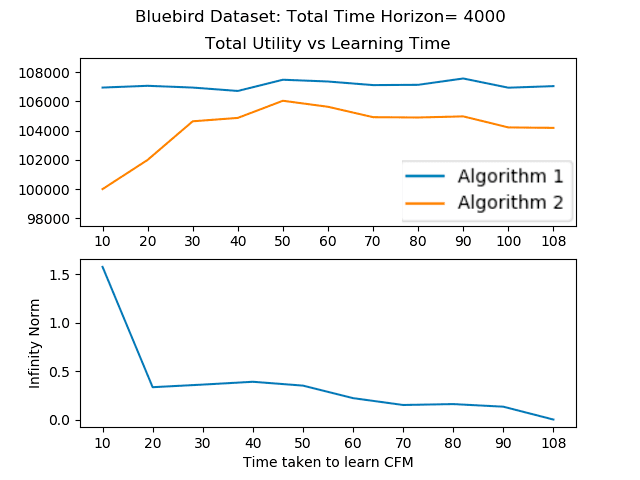}
  \caption{Utility and accuracy with BIRD data set v/s $T_L$.}
  \label{fig:bird}
\end{figure}
\begin{figure}
  \centering
  \includegraphics[width=.9\linewidth]{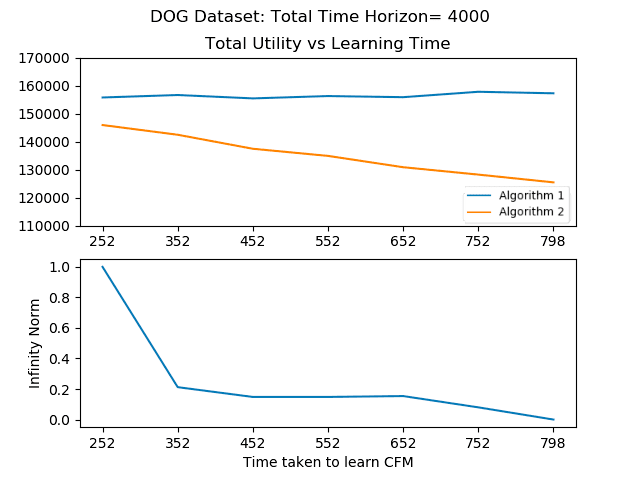}
  \caption{Utility and accuracy with DOG data set v/s $T_L$.}
  \label{fig:dog}
\end{figure}

{\bf Real Data Sets:}Next, we use real two datasets, the Bird dataset \cite{bird} that has $M= 39$ classifiers/workers, and $|S|=108$ samples in total, and the DOG dataset \cite{dog} for which 
we take $M= 90$ classifiers and $|S|=798$ samples in total. 
We modified the multi-class DOG dataset to a binary dataset by clubbing classes 
$1,2$ as class $1$ and classes $3,4$ as class $2$. 
Similar to Fig. \ref{fig:synth}, we first plot the utilities of  Algorithm \ref{alg:greedy-genie} and Algorithm  \ref{alg:realgreedy} for a fixed value of time horizon $T=4000$, together with the accuracy of learnt confusion matrices in Figs. \ref{fig:bird} and Fig. \ref{fig:dog}, for the Bird and the DOG datasets, respectively.
To simulate genie access of $p_m$ for Algorithm \ref{alg:greedy-genie}, the confusion matrix 
learnt after using all the samples is taken as the input at the start itself. 
In Figs. \ref{fig:birdregret} and  \ref{fig:dogregret}, we plot the regret (normalized) and the competitive ratio between Algorithm \ref{alg:greedy-genie} and Algorithm \ref{alg:realgreedy}, as a function of time horizon $T$, where for each value of $T$, the best learning interval $T_L$ is found. More detailed results can be found in the Appendix.

%
%

\begin{figure}[!h]
\centering
\includegraphics[width= 0.65\linewidth]{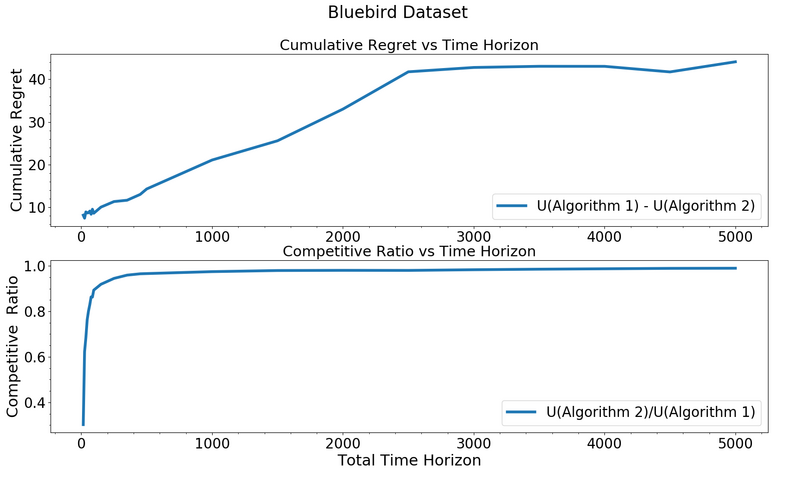}
\caption{Regret and competitive ratio for Algorithm \ref{alg:realgreedy} and Algorithm \ref{alg:greedy-genie} with  Bird data set }
\label{fig:birdregret}
\end{figure}

\begin{figure}[!h]
\centering
\includegraphics[height= 0.4\linewidth]{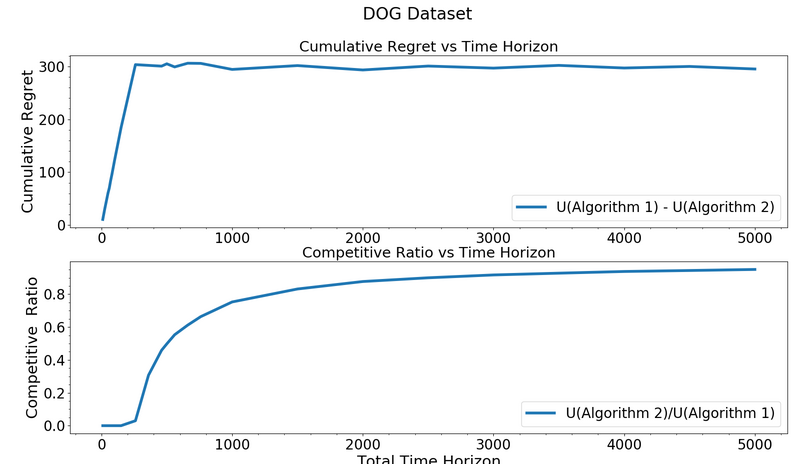}
\caption{Regret and competitive ratio for Algorithm \ref{alg:realgreedy} and Algorithm \ref{alg:greedy-genie} with  DOG data set }
\label{fig:dogregret}
\end{figure}


\end{document}